\documentclass{article}

\usepackage{PRIMEarxiv}

\usepackage[utf8]{inputenc} 
\usepackage[T1]{fontenc}    
\usepackage{hyperref}       
\usepackage{url}            
\usepackage{booktabs}       
\usepackage{amsfonts}       
\usepackage{nicefrac}       
\usepackage{microtype}      
\usepackage{lipsum}
\usepackage{fancyhdr}       
\usepackage{graphicx}       
\graphicspath{{media/}}     
\usepackage{times}
\usepackage{soul}
\usepackage{url}
\usepackage[small]{caption}
\usepackage{graphicx}
\usepackage{amsmath}
\usepackage{amsthm}
\usepackage{booktabs}
\usepackage{algorithm}
\usepackage{algorithmic}
\usepackage[switch]{lineno}
\usepackage{amsmath}
\usepackage{amsthm}
\usepackage{amssymb}
\usepackage{booktabs}
\usepackage{algorithm}
\usepackage{algorithmic}
\usepackage{float}
\usepackage{amssymb}
\usepackage{color}
\usepackage{multicol}
\usepackage{multirow}
\title{ESP-zero: Unsupervised enhancement of zero-shot 
classification for Extremely Sparse Point cloud
}

\author{
  Jiayi Han, Weibo Zheng, Xiangguo Zhou, Daisen Wei$^{\dag}$ \\
  Inspur Gensoft Co. Ltd., Inspur Group Co. Ltd. \\
  Jinan\\
  \texttt{\{hanjiayi, zhouxg, zhengwb, weids\}@inspur.com} \\
   \And
  Zidi Cao \\
  Zhejiang University \\
  Hangzhou\\
  \And
  Yuanfang Zhang$^{\dag}$ \\
  School of Computer Science \\ Nanjing University of Information Science and Technology \\
  Nanjing \\
  \texttt{\{zyf.robinzhang\}@gmail.com}  \\
  \And
  Xiangjian He$^{\dag}$ \\
  Computer Vision and Intelligent Perception Lab\\ University of Nottingham \\
  Ningbo \\
  \texttt{\{sean.he\}@nottingham.edu.cn}
  \thanks{$^{\dag}$: Corresponding author} 
}

\begin{document}
\maketitle

\begin{abstract}
In recent years, zero-shot learning has attracted the focus of many researchers, due to its flexibility and generality. Many approaches have been proposed to achieve the zero-shot classification of the point clouds for 3D object understanding, following the schema of CLIP. However, in the real world, the point clouds could be extremely sparse, dramatically limiting the effectiveness of the 3D point cloud encoders, and resulting in the misalignment of point cloud features and text embeddings. To the point cloud encoders to fit the extremely sparse point clouds without re-running the pre-training procedure which could be time-consuming and expensive, in this work, we propose an unsupervised model adaptation approach to enhance the point cloud encoder for the extremely sparse point clouds. We propose a novel fused-cross attention layer that expands the pre-trained self-attention layer with additional learnable tokens and attention blocks, which effectively modifies the point cloud features while maintaining the alignment between point cloud features and text embeddings. We also propose a complementary learning-based self-distillation schema that encourages the modified features to be pulled apart from the irrelevant text embeddings without overfitting the feature space to the observed text embeddings. Extensive experiments demonstrate that the proposed approach effectively increases the zero-shot capability on extremely sparse point clouds, and overwhelms other state-of-the-art model adaptation approaches.
\end{abstract}


\section{Introduction}

\begin{figure*}[htbp]
    \centering
    \includegraphics[width=.65\linewidth]{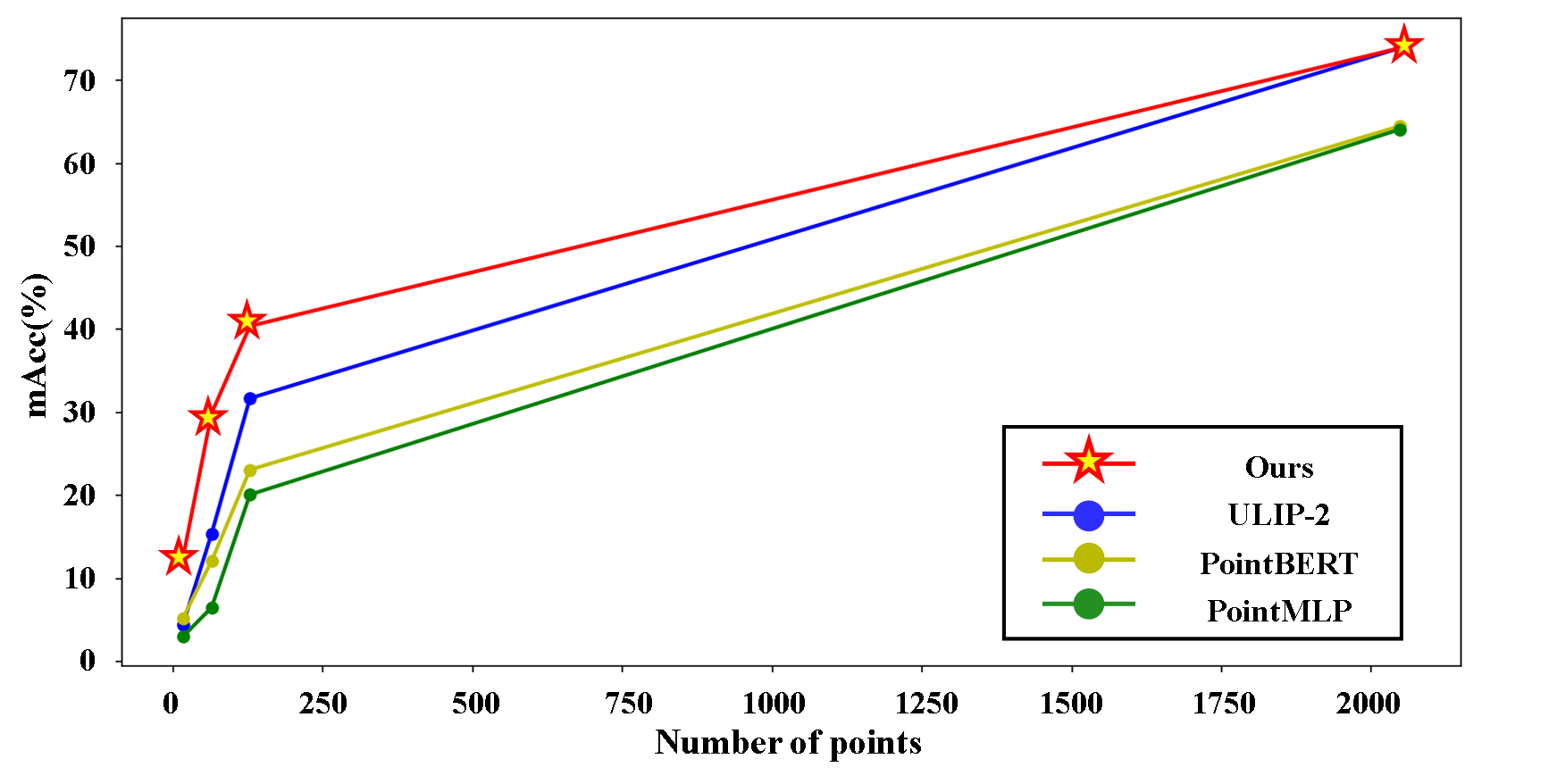}
    \caption{Zero-shot classification accuracy on the ModelNet-40 dataset. Our approach dramatically increases the zero-shot capability on the extremely sparse point clouds.}
    \label{fig:comp}
\end{figure*}

3D point cloud processing with deep-learning approaches has been explored deeply in recent years, due to its growing applications in VR, AR, autonomous driving, embodied intelligence, and so on. With the improvements in self-attention techniques, the performance of 3D point cloud processing has significantly increased. 

Despite the great success of 3D point cloud processing, researchers mainly focus on dense point clouds, while the processing of extremely sparse point clouds has not been fully explored, especially in zero-shot settings. As the density of points significantly influences the shape representation of the input object, the models trained with dense point clouds may lack generalization ability on those extremely sparse point clouds. 

Inspired by the pre-training schema in NLP tasks, some 3D object processing approaches with the pre-training step involve the training procedure of sparse point clouds. For example, PointBERT \cite{PointBERT} groups the input point cloud into clusters via furthest point sampling (FPS), encodes each cluster into an embedding, masks $25\%\sim 45\%$ clusters out of the embeddings, and then recovers those masked embeddings.

This training approach results in the training on the sparse point clouds. Similarly, I2P-MAE \cite{I2P-MAE} also proposes to mask out $80\%$ of points during the pre-training procedure. However, with an initial number of 8096 points, the down-sampled point cloud still has more than 1k points which is still very dense and may not be applicable in real usages.

A potential solution for enhancing sparse point cloud classification is to complete the sparse point clouds, named point completion. However, researchers of point cloud completion mainly focus on completing the sparse point clouds with approximately 2k points to 16k points \cite{lin2023hyperbolic,PmpNet,Pmpnet++,SeedFormer}. Some of the point-completion approaches validated their performance on datasets with more sparse samples, for example, the KITTI dataset \cite{KITTI}. However, these methods are only validated on a limited number of categories (usually only ''vehicle'' is validated). 

\begin{figure*}[htbp]
    \centering
    \includegraphics[width=0.9\linewidth]{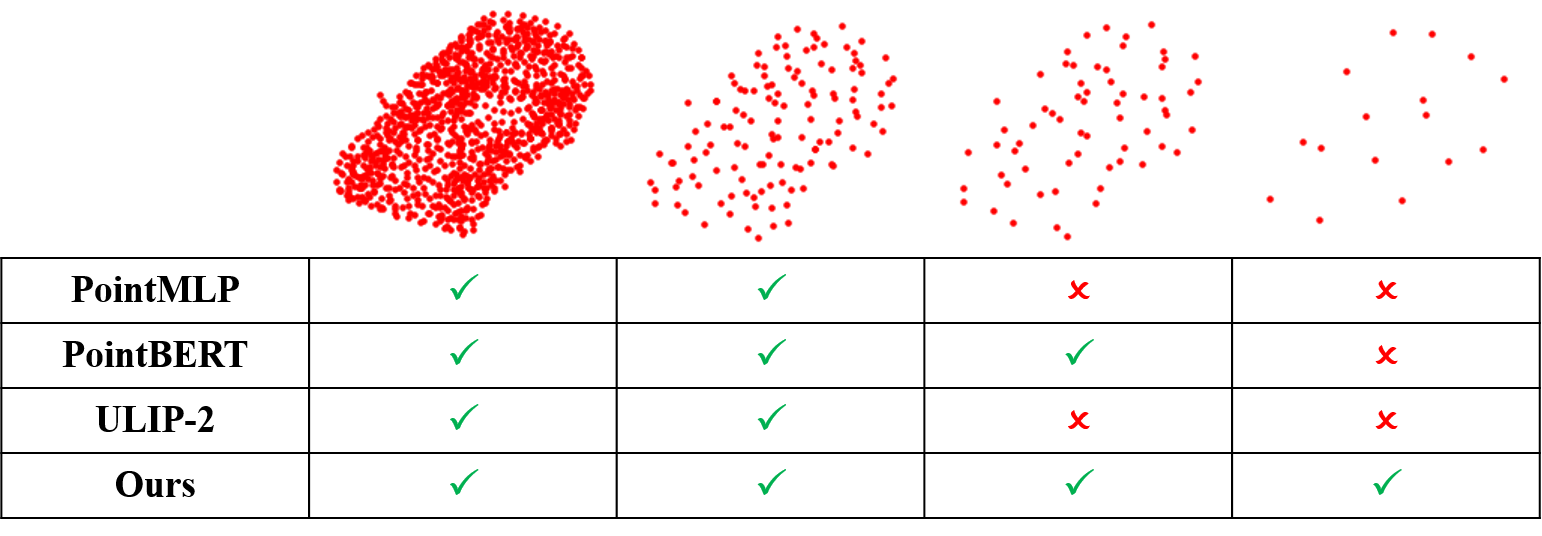}
    \caption{The proposed approach enhances the zero-shot ability on the extremely sparse point clouds.}
    \label{fig:fig1}
\end{figure*}

To enhance the zero-shot classification capability on the extremely sparse point clouds, in this work, we propose a novel self-distillation approach that leverages the information of the dense point cloud and the text embedding of the observed categories. 
The proposed approach is based on a well-trained point cloud transformer, whose latent space is aligned with the latent space of text embeddings. 
For the alignment of point cloud feature space and text embedding space could be reserved, we introduce an additive tuning strategy. 
We freeze the weights of the pre-trained point cloud transformer and attach fused cross-attention (FCA) layers to it for model optimization. Each FCA layer consists of a batch of learnable tokens, a learnable self-attention (SA) block, and a frozen self-attention (SA) block that belongs to the pre-trained network. 
Following the cross-attention mechanism, the learnable tokens are first fed into the learnable SA block. Inspired by VPT \cite{VPT}, the modified tokens are concatenated to the encoded point tokens. The frozen SA block further encodes the merged tokens, and the learnable tokens are discarded in the output. Despite the small number of learnable tokens, FCA could effectively enhance the pre-trained model.

We also propose a novel complementary learning-based self-distillation approach to optimize the modified model for both visible and invisible categories in the training set. Different from normal distillation approaches that utilize the pseudo labels or modified output distributions of the input, we adopt complementary labels to specify the categories to which the input does not belong, and suppress the similarity of the 3D object representations and the text embeddings of the complementary labels. This approach allows the encoded sparse representation to be pulled apart from the unmatched text embeddings rather than fitting the most similar text embedding, which reduces the risk of overfitting the observed text embeddings during training. Our main contributions could be listed as follows.
\begin{enumerate}
\item To the best of our knowledge, this work proposes to enhance the zero-shot classification of extremely sparse point clouds for the first time.
\item We propose a fused cross-attention layer that introduces the refinement of frozen self-attention blocks, which effectively modifies the encoded representation space of the pre-trained model while maintaining its zero-shot capability.
\item We propose a complimentary learning-based self-distillation approach that pulls the sparse point cloud representation away from the unmatched label text embeddings, which decreases the potential overfitting.
\end{enumerate}
\section{Related works}
\subsection{Point cloud processing}
Since the success of PointNet \cite{PointNet}, processing 3D objects in the form of point clouds has become a natural solution. PointNet++ \cite{pointnet++} introduces the grouping layers in point cloud processing, which allows the deep models to leverage the neighboring information like convolution networks. Afterward, many researchers propose to utilize well-designed kernels to improve their performances. For example, KP-conv \cite{KPConv} utilizes a spherical neighborhood and a kernel function to determine the weight of each neighboring point during convolution. Edge-conv \cite{EdgeConv} utilizes the relative position to fetch the integrating weight of neighboring points. \cite{hao2022cascaded,huang2023edge} leverage local geometry for point cloud classification. \cite{zhang2022multi} proposes multi-scale FPS to fuse point cloud features. \cite{song2020cnn} transforms points into a Hough space, and utilizes CNN-based networks to encode the points. Augmentations are also beneficial for point cloud processing \cite{meng2024new}. 

In recent years, many approaches have introduced self-attention blocks to point cloud processing. Point Transformer \cite{PointTransformer} and Point Cloud Transformer \cite{PCT} introduce the self-attention mechanism to point cloud processing for the first time. They utilize similar structures like PointNet++ and modify the feature aggregation with self-attention layers. Meanwhile, YOGO \cite{YOGO} proposes to group and embed the points once that first group and encode multiple sub-structures of the point cloud into embeddings, then calculate the cross-attention of grouped embeddings and all the points. This strategy effectively reduces the cost of SA-based point cloud processors but sacrifices some precision. To decrease the cost of self-attention in point cloud processing, PointFormer \cite{PointFormer} proposes to utilize Linformer to replace the standard self-attention mechanism. SD-SA \cite{SD-SA} validates the efficient self-attention mechanisms for the point cloud transformer and proposes to modify self-attention with skeleton decomposition to reduce the computational cost of the point cloud transformer.

Inspired by NLP tasks, PointBERT \cite{PointBERT} and I2P-MAE \cite{I2P-MAE} introduce BERT-styled pre-training in 3D processing and achieve great success. They mask some tokens of the point cloud and train to recover those masked tokens. The fine-tuning procedure is followed to fetch the downstream capabilities. \cite{mei2024unsupervised} utilizes a novel pipeline that leverages neural rendering and 2D images to align features of point clouds and images for effective model pre-training. 

\subsection{Zero-shot classification of point clouds}
Inspired by the success of CLIP \cite{CLIP}, many approaches are proposed to classify point clouds in a zero-shot manner. PointCLIP \cite{PointCLIP} directly renders the point clouds to depth images and adopts a pre-trained CLIP model to classify the rendered images with prompts like “this is the depth map of ``$\{category\}$''. CLIP2Point \cite{clip2point} proposes to render the initial 3D mesh and point cloud into images and depth maps, encode them by pre-trained CLIP models, and fine-tune the model to align their features. \cite{cheraghian2022zero} aligns the seen semantics and point cloud features, and leverages the unlabeled object to address the downstream issues like domain adaption. ULIP and ULIP-2 \cite{ULIP,ULIP-2} further introduce the text embeddings in the training phase. The point cloud features are aligned to both CLIP features and corresponding text embeddings of the ground truth labels.

\subsection{Point cloud completion}
Point completion network (PCN) \cite{PCN} is one of the most classical point cloud completion approaches, which introduces both coarse and fine-grained supervision during training. Many approaches pay attention to refining the details of the completed point clouds. PMP-Net \cite{PmpNet} proposed to iteratively refine the sparse point cloud. After each refinement step, the modified point cloud is further refined, until it reaches the maximum refinement steps. PMP-Net++ \cite{Pmpnet++} further enhances the PMP-Net with the self-attention mechanism. SeedFormer \cite{SeedFormer} introduces patch seeds as shape representation which leverages both global information (of the partial point cloud) and local geometry. Similarly, \cite{zhang2023learning} also leverages the geometry transformation to recover the missing point. USSPA \cite{ma2023symmetric} introduces a shape-preserving autoencoder for point cloud completion without supervision. Adapted loss functions are also introduced to enhance the point cloud completion \cite{lin2023hyperbolic}.
\section{Method}

\subsection{Overall Architecture}
\begin{figure*}
    \centering
    \includegraphics[width=.95\linewidth]{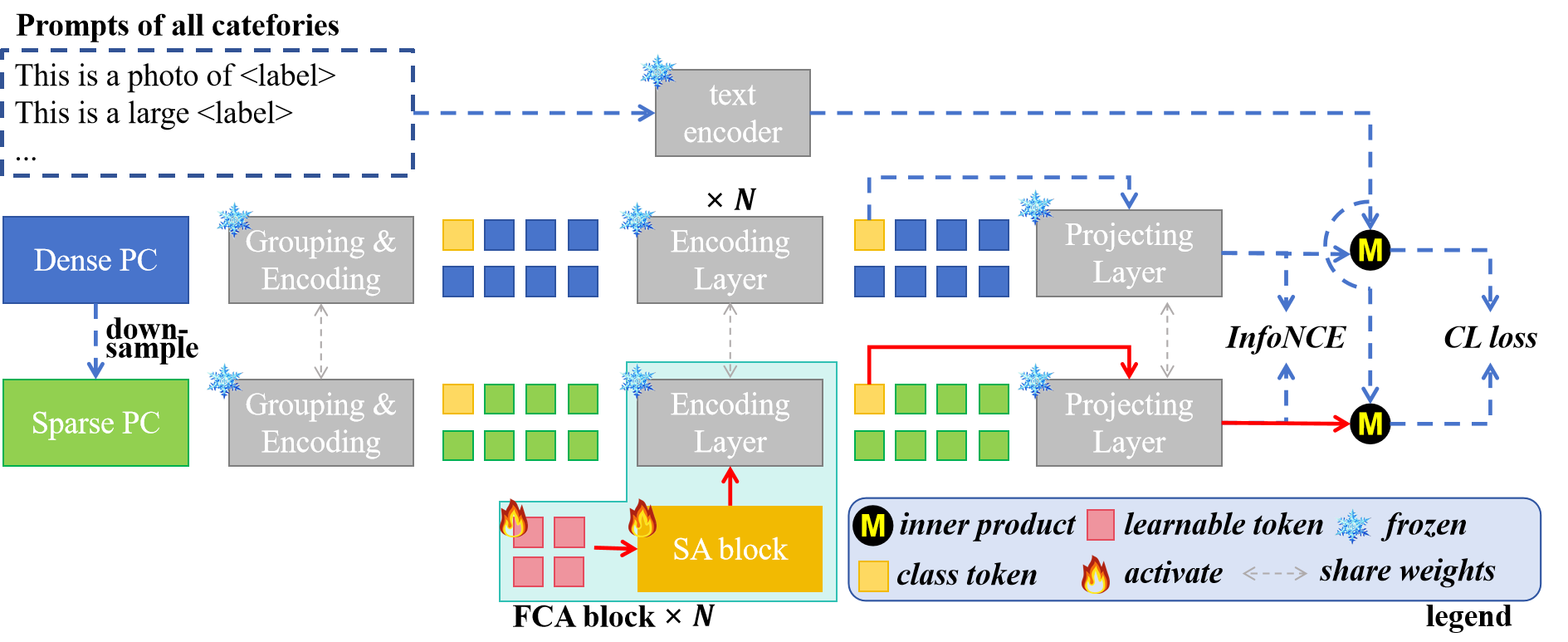}
    \caption{The overall framework of the proposed approach. The Dense point cloud is down-sampled to a sparse point cloud, grouped by KNN, and encoded to point cloud tokens. The initial model is modified with a trainable FCA block when processing the tokens of the sparse point cloud. We then distill the model with infoNCE loss and CL loss with the assistance of text embeddings.}
    \label{fig:overall}
\end{figure*}
As depicted in Fig. 3, the overall framework consists of a pre-trained point cloud transformer with grouping and encoding layers, encoding layers of the point embeddings, and a final projecting layer that projects the point cloud representation to the latent space shared with text embeddings. The pre-trained network could be seen as a teacher model. 
To modify the pre-trained model, we expand the pre-trained encoding layers to an FCA block. A trainable FCA block consists of additional learnable tokens, an SA block, and the corresponding frozen encoding layer. By modifying the trainable parts in the FCA block, the model could be enhanced for extremely sparse point clouds. 

During training, the dense point cloud is down-sampled to a sparse point cloud. The dense point cloud is directly encoded with the pre-trained model to obtain a standard representation. The model modified by the FCA block encodes the sparse point cloud. The inner product of text embeddings and the dense representations is utilized to fetch the pseudo supervision for the sparse representation to learn from. During testing, the text embedding with the largest similarity to the sparse representation indicates the final prediction. Note that since the pre-trained weights are fixed, and the FCA block could be directly rolled back to the initial encoding layer, therefore we could avoid catastrophic forgetting.

\subsection{Fused Cross-attention (FCA)}

The pre-trained point-cloud transformer is well-aligned with text embedding, which is crucial to zero-shot classification. To fine-tune the pre-trained model without eliminating its zero-shot capability, we introduce fused cross-attention to each transformer block of the point-cloud transformer. 

For each transformer block, we add $m$ learnable tokens which are randomly initialized. We first forward the learnable tokens with a self-attention block, then fuse them to the encoded point cloud tokens. Specifically, denote the encoded point cloud tokens as $T=\{t_i\}_{i=1}^{n}$ and the learnable tokens as $P=\{p_j\}_{j=1}^{m}$. The learnable tokens are first processed as in Eq.~\ref{eq:ca1}:
\begin{equation}
\left\{
    \begin{array}{l}
        \hat{P} = \frac{PW_q(PW_k)^T}{\lambda}PW_v\\
        P' = \hat{P}W_P,
    \end{array}
    \right.
    \label{eq:ca1}
\end{equation}
in which SA and FFN represent the self-attention and the feed-forward network, respectively. $P'$ is then fused with $T$ via self-attention as follows:
\begin{equation}
\left\{
    \begin{array}{l}
        \hat{T} = \frac{[T;P']W_q([W;P']W_k)^T}{\lambda}[T;P']W_v\\
        T' = \hat{T}W_T,
    \end{array}
    \right.
\end{equation}
After the fusion procedure, the modified point tokens are fed forward, while the learnable tokens are discarded so that the total number of tokens is consistent. The final output of each FCA layer is as $T'_{0:\vert T\vert}$.

\begin{figure}
    \centering
    \includegraphics[width=.6\linewidth]{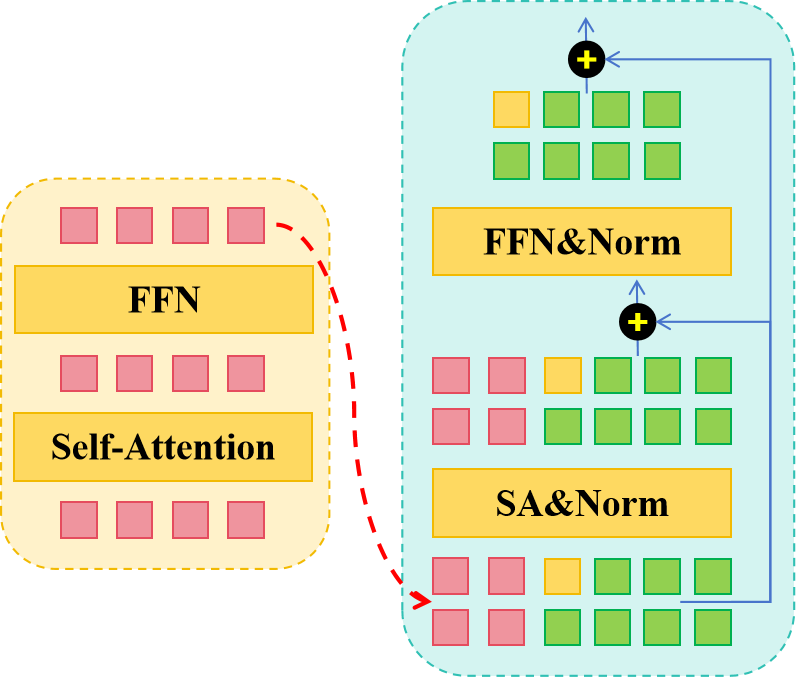}
    \caption{The structure of FCA. The learnable tokens are firstly processed by an SA block, then combined with the encoded point cloud tokens, pass through an encoder block, and only output the point cloud tokens.}
    \label{fig:FCA}
\end{figure}

\subsection{Self-distillation with complementary learning}
\begin{figure}
    \centering
    \includegraphics[width=.6\linewidth]{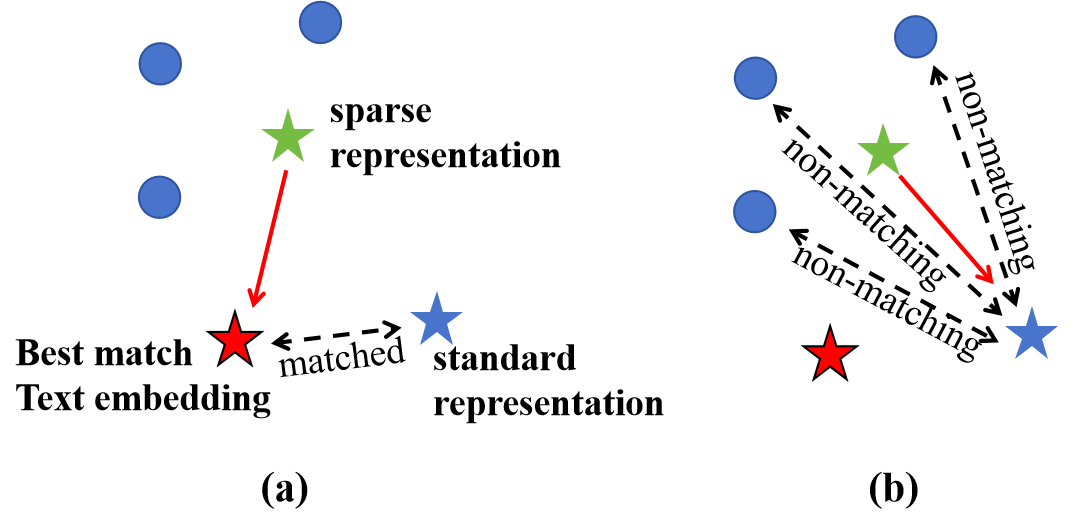}
    \caption{The difference between pseudo label and complementary label. Via pseudo labeling (a), the sparse representation is encouraged to be aligned with the text embedding of the pseudo label. On the contrary, complementary learning (b) encourages the sparse representation to be apart from the unmatched text embeddings. }
    \label{fig:plcl}
\end{figure}
To enhance the classification ability of the pre-trained point cloud transformer, we propose a self-distillation schema to optimize the learnable parameters (the learnable tokens, $q,k,v$ projections, and FFN) in FCA. 
The pre-trained point cloud transformer without FCA is utilized to encode the dense point cloud (with 2048 points) to form its standard representation which is well-aligned to the text embedding of the prompt template of the corresponding category. The text embedding could be obtained by the corresponding text encoder, in our case, a text transformer aligned with the pre-trained point cloud transformer.

\paragraph{Pseudo label for Self-distillation}
Denote the standard representation as $R\in\mathbb{R}^{1,C}$, and the text embeddings of the prompt templates (as shown in Fig.~\ref{fig:overall}) of involved categories as $E\in\mathbb{R}^{N,C}$, in which $N, C$ represent the number of involved categories and the dimensionality of the latent space, respectively.
The complementary labels of the input objects could be obtained by the similarity of the standard representation and text embeddings. We first compute the similarity of the standard representation and the text embeddings:
\begin{equation}
    \begin{array}{l}
        q = \frac{ER^T}{||E||_2||R||_2}.
    \end{array}
\end{equation}
 A direct optimization approach could be utilizing the pseudo labels. For a given input $x$, its pseudo label could be formulated as follows:
\begin{equation}
(\hat{y}|x)_i = 
\left\{
    \begin{array}{l}
        1, i=\mathop{\text{argmax}}\limits_i q_i\\
        0, otherwise
    \end{array}
\right.
\end{equation}

Then we obtain the sparse representations. In this work, we down-sample the dense point cloud according to the uniform distribution. We then encode the down-sampled point cloud with the modified model (with FCA) to obtain the sparse representations. 
The cross-entropy loss could be formulated as in Eq.~\ref{eq:plloss}:
\begin{equation}
    \begin{array}{l}
        loss_{pl} = -\sum\limits_i (\hat{y}|x)_i \text{log}\big(\text{Softmax}(\hat{q}/\tau)_i\big)
    \end{array}
    \label{eq:plloss}
\end{equation}
By optimizing Eq.~\ref{eq:plloss}, the object representation $R_{|x}$ could be modified to match the most probable text embedding. 
Although utilizing pseudo labels could be a direct self-distillation approach since the matched embedding could be seen as an approximation of the standard representation, this process might result in the model overfitting on the training text embeddings, and decrease the zero-shot ability on the unseen categories. 

\paragraph{Complementary learning for dense-to-sparse self-distillation} Different from learning with pseudo labels which directly minimize the distance between the sparse representation and the matched text embeddings, complementary learning aims to learn from the labels that an input does not belong to, namely ``complementary labels''. Complementary learning allows the model to pull the sparse representation apart from the unmatched text embeddings. ECL\cite{zeng2024rethinking} proves that the complementary labels are more accurate and could provide effective information in unsupervised domain adaptation, compared with pseudo labels. 
To enhance the zero-shot ability of the pre-trained model, instead of utilizing the pseudo labels, we adopt a complementary learning-based approach to leverage the information of labels without overfitting the observed ones. Complementary learning fetches the most ``improbable'' categories and fine-tunes the model to push the object representations apart from those unmatched text embeddings. In this setting, the representations are not matched to a certain embedding.
We select the improbable (or negative) categories based on $q$. The $k$ categories with the smallest similarities to the standard representation are set to negative categories. 
Denote the inner product of the sparse representations and text embedding to $\hat{q}$.
The loss function could be formulated as follows:
\begin{equation}
\begin{array}{l}
    loss_{CL} = -\sum\limits_{i\in C^-}\text{log}(1-\text{Softmax}(\hat{q}/\tau)_i)
\end{array}
\end{equation}
in which $C^-$ represents the set of negative categories and $\tau$ represents temperature.

\paragraph{Total loss}
Apart from the complementary loss, we also introduce infoNCE loss \cite{InfoNCE} to align the sparse representation to the standard representation, which could be formulated as Eq.~\ref{eq:infonce}:
\begin{equation}
    \begin{array}{l}
    loss_{sd} = 
        \frac{\text{exp}(R_i^TR_i^s/\tau)}{\sum_{j\neq i}\text{exp}(R_j^TR^s_i/\tau)},
    \end{array}
    \label{eq:infonce}
\end{equation}
in which $R_i^s$ represents the sparse representation of the $i_{th}$ object. The total loss could be formulated as follows:
\begin{equation}
    \begin{array}{l}
        loss_{total} = \lambda loss_{sd} + loss_{CL},
    \end{array}
\end{equation}
where $\lambda$ is the balance coefficient.
\section{Experiments}
\subsection{Implementation details}
In this work, we validate the proposed approach on two benchmark datasets: ModelNet40\cite{ModelNet40} and PartNet\cite{PartNet}. ModelNet40 dataset contains 12,311 objects (9843 for training) from 40 categories. PartNet dataset contains 26,671 objects from 24 categories. They share nine categories, and the others are unique to each other. All models are trained on a Nvidia-4090 GPU within 16 epochs. We adopt a cross-validation schema that trains the model on one dataset and validates it on the other. We show the result for the ``Unseen'' categories (the unique categories of each dataset). We set the number of learnable tokens to 12, and the coefficient $\lambda$ to 0.2.

\subsection{Comparison with SOTA approaches}
We compare the proposed approach with other zero-shot learners and unsupervised model adaption approaches. 
Note that for PL, Tent\cite{Tent} and USKD-PL\cite{USKD}, we utilize the prediction of 2048 points to obtain the pseudo label or label distribution, for a fair comparison with the proposed approach. However, due to the nature of RPL \cite{RPL}, we could not involve the standard representation during adaptation, so only the sparse representation is utilized. 
In Tab.~\ref{tab:unseen}, we demonstrate the model performance on the \textbf{unseen} classes that are unique to each dataset. The result shows that our approach dramatically increases the classification of the unseen categories, in an average of $8.1\%$ on the PartNet dataset, and $8.7\%$ on the ModelNet40 dataset, which demonstrates the effectiveness of our method.

\begin{table}[htbp]
  \centering
  \caption{Zero-shot classification accuracy of \textbf{UnSeen} categories on ModelNet40 and PartNet dataset.}
    \begin{tabular}{l|cccc|cccc}
    \toprule
    Dataset & \multicolumn{4}{c|}{M40$\rightarrow$Partnet} & \multicolumn{4}{c}{PartNet$\rightarrow$M40} \\
    \midrule
    Num Points & 128   & 64    & 16    & Mean  & 128   & 64    & 16    & Mean \\
    \midrule
    PointBERT\cite{PointBERT} & 45.2  & 17.0  & 0.2   & 20.8  & 23.1  & 12.1  & 5.2   & 13.5  \\
    PointMLP\cite{PointMLP} & 40.3  & 10.8  & 0.0   & 17.0  & 20.1  & 6.5   & 3.0   & 9.9  \\
    ULIP-2\cite{ULIP-2} & 33.3  & 17.6  & 7.8   & 19.6  & 31.7  & 15.4  & 4.4   & 17.2  \\
    \midrule
    Pseudo Label & 38.9  & 29.5  & 13.8  & 27.4  & 25.8  & 15.7  & 3.6   & 15.0  \\
    TENT\cite{Tent}  & 40.1  & 22.4  & 9.8   & 24.1  & 30.6  & 16.9  & 3.4   & 17.0  \\
    RPL\cite{RPL}   & 30.6  & 16.8  & 11.1  & 19.5  & 31.0  & 15.6  & 3.1   & 16.6  \\
    USKD\cite{USKD} & 27.2  & 26.1  & 19.7  & 24.3  & 25.5  & 17.5	& 6.4	& 16.5  \\

    USKD-PL\cite{USKD}  & 27.2  & 16.4  & 3.3  & 15.6  & 31.8 	& 19.2 & 5.7 & 18.9\\
    Ours  & \textcolor[rgb]{ 1,  0,  0}{\textbf{45.8}} & \textcolor[rgb]{ 1,  0,  0}{\textbf{39.9}} & \textcolor[rgb]{ 1,  0,  0}{\textbf{20.8}} & \textcolor[rgb]{ 1,  0,  0}{\textbf{35.5 }} & \textcolor[rgb]{ 1,  0,  0}{\textbf{40.4 }} & \textcolor[rgb]{ 1,  0,  0}{\textbf{29.9 }} & \textcolor[rgb]{ 1,  0,  0}{\textbf{12.4 }} & \textcolor[rgb]{ 1,  0,  0}{\textbf{27.6 }} \\
    \bottomrule
    \end{tabular}%
  \label{tab:unseen}%
\end{table}%

\subsection{Point cloud completion may not be sufficient for extremely sparse point cloud zero-shot classification}
It is a direct approach to improving the classification performance by point cloud completion. In this work, we validate two popular point cloud completion methods: PCN \cite{PCN} and SeedFormer \cite{SeedFormer}. For the PCN, due to the flexibility of PCN architecture, we modify the PCN network by changing the coarse output to 256 points and the fine-grained output to 2048 points and training the modified PCN in both end-to-end and independent manners. The end-to-end manner means that the modified PCN is trained along with the pre-trained ULIP model.
For the independent manner, we first train point completion independently, then merge the modified PCN and the ULIP model. SeedFormer leverages grouping layers which are introduced in PointNet++. As there are many fixed parameters that could influence the model's performance, we maintain its initial architecture and utilize its pre-trained model. Since the SeedFormer holds the limitation of 256 input points, we simply repeat the point cloud to meet its limitation. The results are shown in Tab.1. 
Seedformer pre-trained with PCN \cite{PCN} dataset lacks generality on the extremely sparse point clouds of the ModelNet40 dataset, thus resulting in a performance decrease. Even if the model is fine-tuned for ModelNet-40, a performance decrease is still observed. Only when the PCN is trained in an end-to-end manner can the performance be increased, which demonstrates that simply applying point cloud completion is not sufficient for extremely sparse point cloud zero-shot classification.
\begin{table}[htbp]
  \centering
  \caption{Validation of point cloud completion for extremely sparse point cloud (128 points) on the FULL ModelNet40 test set.}
    \begin{tabular}{c|c|c}
    \toprule
    Model & Training strategy & Acc(\%) \\
    \midrule
    SeedFormer\cite{SeedFormer} & pre-trained on PCN & 5.0 \\
    PCN-modified\cite{PCN} & two-stage & 20.9 \\
    PCN-modified\cite{PCN} & end-to-end & 37.2 \\
    \midrule
    No Completion & /     & 35.4 \\
    \bottomrule
    \end{tabular}%
  \label{tab:pcn}%
\end{table}%

\subsection{Ablation study}

\paragraph{Validation of the proposed modules}
We first validate the effectiveness of the proposed modules on both ModelNet40 and PartNet. The result is shown in Tab.~\ref{tab:mainabl}. 
``CL'' demonstrates the complementary learning, and ``CA'' demonstrate the cross-attention.
Note that the baseline represents the pre-trained ULIP-2 backbone and the model trained only by InfoNCE needs an additional MLP head following the model. The result demonstrates that by introducing the learnable tokens, the model performance is significantly increased. By adopting the cross attention in FCA, the model is further improved, with an average of $1.3\%$ performance gain. The complementary loss dramatically increases the model's performance, on an average of $5.6\%$, which demonstrates the effectiveness of the proposed modules.

\begin{table}[htbp]
  \centering
  \caption{The ablation study of the proposed modules, validated on the PartNet dataset. Note that the baseline (no modules are involved) represents ULIP-2. When only InfoNCE is adopted, a learnable MLP is followed to the pre-trained model.}
    \begin{tabular}{cc|cc|ccc}
    \toprule
    \multicolumn{2}{c|}{Self-distillation} & \multicolumn{2}{c|}{FCA} & \multicolumn{3}{c}{Accuracy} \\
    \midrule
    InfoNCE & CL    & tokens & CA   & N=128 & N=64  & N=16 \\
    \midrule
    $\times$ & $\times$ & $\times$ & $\times$ & 40.6$\%$      & 24.8$\%$      & 9.0$\%$ \\
    $\checkmark$ & $\times$ & $\times$ & $\times$ & 31.4$\%$      & 22.2$\%$      & 6.7$\%$ \\
    $\checkmark$ & $\times$ & $\checkmark$ & $\times$ & 44.2$\%$      & 39.6$\%$      & 25.9$\%$ \\
    $\checkmark$ & $\times$ & $\checkmark$ & $\checkmark$ & 45.4$\%$      & 40.5$\%$      & 27.5$\%$ \\
    $\checkmark$ & $\checkmark$ & $\checkmark$ & $\checkmark$ & 53.2$\%$      & 47.9$\%$      & 29.2$\%$ \\
    \bottomrule
    \end{tabular}%
  \label{tab:mainabl}%
\end{table}%

\paragraph{Further validation of point cloud completion}
To further validate whether point completion is beneficial to super sparse point clouds, we also validate the combination of PCN+learnable tokens+InfoNCE and show the result in Tab.~\ref{tab:fpcn}. The PCN results in a crucial performance decrease, especially with a short number of points. This result demonstrates that the PCN is not able to perform robust point completion, and might overfit on the training samples, which leads to the loss of generality in zero-shot classification.

\begin{table}[htbp]
    \centering
    \caption{Further validation of point cloud completion for zero-shot super sparse point cloud classification on PartNet dataset.}
    \begin{tabular}{c|c|c|c}
    \toprule
       -        & N=128  & N=64   & N=16\\
    \midrule
       w PCN    & 49.7$\%$ & 37.0$\%$ & 15.1$\%$ \\
       w/o PCN  & 53.2$\%$ & 47.9$\%$ & 30.2$\%$ \\
    \bottomrule
    \end{tabular}
    
    \label{tab:fpcn}
\end{table}

\paragraph{Number of learnable tokens}
We also validate the influence of the number of learnable tokens, without the cross attention, on the \textbf{UnSeen} categories of PartNet. As shown in Tab.~\ref{tab:ntokens}, increasing the number of learnable tokens benefits the performance, but the improvement becomes marginal with the number getting larger.
\begin{table}[H]
  \centering
  \caption{The influence of the number of learnable tokens. Increasing the number of learnable tokens benefits the model performance, but the improvement is getting marginal with $N$ getting large.}
    \begin{tabular}{c|ccc|c}
    \toprule
    \multirow{2}[2]{*}{Num of Tokens} & \multicolumn{3}{c|}{Num of Points} & \multirow{2}[2]{*}{Mean} \\
          & 128   & 64    & 16    &  \\
    \midrule
    4     & 32.0  & 21.9  & 7.3   & 20.4  \\
    8     & 36.6  & 31.3  & 18.6  & 28.8  \\
    12    & 43.0  & 35.1  & 20.8  & 33.0  \\
    24    & 41.2  & 36.6  & 25.6  & 34.5 \\
    \bottomrule
    \end{tabular}%
  \label{tab:ntokens}%
\end{table}%

\paragraph{Down-sampling strategies}
Except for the random down-sampling, we also validate the model performance of the KNN down-sampling strategy. This strategy selects a random center point and then samples the K nearest points to the center point. The result is shown in Tab.~\ref{tab:strategy}, which demonstrates that the proposed approach could generalize on different down-sampling strategies.
\begin{table}[H]
    \centering
    \caption{The influence of different down-sampling strategies during the test phase on the ModelNet40 dataset. Compared with the ULIP-2 and its enhancement via pseudo labels, the proposed approach achieves a significant improvement.}
    \begin{tabular}{c|c|c|c|c}
    \toprule
       Model   & N=128  & N=64  & N=16 & Mean\\
    \midrule
       ULIP-2  & 27.7   & 13.2  & 3.3  & 14.7\\ 
       PL      & 29.9   & 17.7  & 4.0  & 17.2\\
       Ours    & 38.0   & 26.5  & 6.6  & 23.7\\
    \bottomrule
    \end{tabular}
    
    \label{tab:strategy}
\end{table}

\section{Conclusion}
In this work, we raise the issue of zero-shot super sparse point cloud classification for the first time and propose a simple yet effective unsupervised training schema that effectively enhances the zero-shot classification ability of super sparse point clouds. We propose a learnable FCA to modify the latent space of the point cloud encoder while maintaining its alignment with text embeddings. We also propose a complementary learning-based self-distillation approach that leverages the training labels without overfitting the training text embeddings. Explicit experiments demonstrate the effectiveness of the proposed approach, which dramatically increases the zero-shot classification performance on the super sparse point clouds.

\bibliographystyle{unsrt}  
\bibliography{references}

\end{document}